\documentclass[sigconf]{acmart}
\AtBeginDocument{%
  }

\setcopyright{acmlicensed}
\copyrightyear{2026}
\acmYear{2026}
\acmDOI{XXXXXXX.XXXXXXX}
\acmConference[KDD '26]{The Undergraduate Consortium at KDD 2026}{August 9--13, 2026}{Republic of Korea}
\acmISBN{979-8-4007-2258-5/2026/08}

\usepackage{amsmath}
\usepackage{algorithm}
\usepackage{algorithmic}
\usepackage{booktabs}
\usepackage{graphicx}
\usepackage{multirow} 
\usepackage{float}   
\usepackage{mathtools}

\newcommand{\best}[1]{\textbf{#1}}
\newcommand{\second}[1]{\underline{#1}}

\allowdisplaybreaks
\begin{document}

\title{SafeECGMatch: Calibration-Aware Joint Frequency and Time Space Semi-Supervised Learning for Open-Set ECG Classification}



\author{Hongkyu Koh}
\email{202200307@hufs.ac.kr}
\affiliation{%
  \institution{Hankuk University of Foreign Studies}
  \city{Yongin}
  \country{Republic of Korea}
}

\author{Ikbeom Jang} 
\authornote{Corresponding author.}
\email{ijang@hufs.ac.kr}
\affiliation{%
  \institution{Hankuk University of Foreign Studies}
  \city{Yongin}
  \country{Republic of Korea}
}

\begin{abstract}
Electrocardiogram (ECG) classification models often suffer from severe label scarcity, making semi-supervised learning (SSL) an attractive strategy for reducing annotation costs. In clinical settings, however, unlabeled pools frequently contain out-of-distribution (OOD) anomalies or diagnostic groups absent from the labeled set. Standard SSL forces incorrect pseudo-labels onto these unseen classes, producing overconfident predictions. 
To address this, we propose SafeECGMatch, a calibration-aware safe SSL framework for single-label ECG classification under label distribution mismatch. 
Methodologically, SafeECGMatch employs a dual-branch architecture extracting time-frequency latent representations via ECG-specific augmentations. Crucially, it dynamically aligns confidence with empirical accuracy through adaptive label smoothing and temperature scaling, calibrating both the multiclass classifier and the OOD detector across temporal and spectral domains. This joint optimization allows trustworthy OOD rejection and reliable pseudo-labeling. Evaluated on the PTB-XL and PhysioNet/CinC
Challenge benchmarks, SafeECGMatch achieves state-of-the-art accuracy and calibration, advancing reliable knowledge discovery in physiological time-series. Code is available at \url{https://github.com/labhai/SafeECGMatch}.
\end{abstract}



\maketitle

\section{Introduction}

Electrocardiography (ECG) is one of the most widely used non-invasive tools for detecting and monitoring cardiovascular disorders in clinical practice. ECG can be used to diagnose conditions such as myocardial infarction, abnormal cardiac rhythms, and angina  \cite{vrints20242024}. In addition, while the diagnosis of heart failure typically requires costly echocardiography, ECG has increasingly been utilized as a powerful screening and prediction tool for identifying patients with latent asymptomatic heart failure at an early stage \cite{attia2019screening}.


With the rapid progress of deep neural networks, automated ECG interpretation systems have achieved increasingly competitive diagnostic performance and have shown strong potential for supporting large-scale screening and clinical decision-making~\cite{wang2024medformer, cai-etal-2025-supreme}. Despite this progress, building reliable ECG classification models still depends on expert annotations, which are expensive, time-consuming, and often difficult to obtain at scale. This bottleneck is particularly severe in medical domains, where labeling requires specialized clinical expertise and strict quality control. Semi-supervised learning (SSL), which jointly leverages a small set of labeled samples and a large pool of unlabeled samples, has therefore emerged as a practical and effective strategy for reducing annotation cost while maintaining strong predictive performance~\cite{sohn2020fixmatch}.

However, the standard SSL assumption that labeled and unlabeled data share the same label space is frequently violated in real-world clinical environments. In practice, unlabeled ECG recordings collected from hospitals or monitoring systems can contain rhythms, morphologies, or patient subgroups that are not represented in the labeled training set. Such a label distribution mismatch introduces out-of-distribution (OOD) samples into the unlabeled data. Applying conventional SSL under this setting can force incorrect pseudo-labels onto unseen classes, and once these errors are fed back into training, they can severely degrade both in-distribution (ID) classification and overall model reliability. To address this issue, Safe SSL, also referred to as open-set SSL, has been proposed to combine semi-supervised classification with OOD detection so that harmful unlabeled outliers can be identified and excluded during learning~\cite{saito2021openmatch,li2023iomatch, wang2024scomatch,he2022safe}.

Although existing Safe SSL methods have made meaningful progress, they remain vulnerable to a persistent problem of modern deep neural networks: overconfidence. In both pseudo-labeling and OOD detection, the model may assign high confidence to erroneous predictions, causing mislabeled unlabeled samples or OOD instances to be incorporated into training with excessive certainty. This failure is particularly problematic in medical applications, where reliability is as important as accuracy \cite{kim2026morcu}. Recent work, such as CaliMatch, has shown that calibration can serve as an effective method by jointly improving the calibration of both the classifier and the OOD detector~\cite{bae2025calimatch}. Nevertheless, CaliMatch has been developed almost exclusively for image recognition benchmarks. In the ECG domain, prior SSL studies like ECGMatch~\cite{zhou2023semi} have demonstrated the promise of leveraging unlabeled signals, but they do not directly address the open-set scenario or explicitly tackle overconfidence. This gap is significant because ECG signals exhibit strong temporal structure, morphology-dependent variation, and clinically meaningful inter-patient heterogeneity, making confidence estimation and distribution shift handling far more challenging than in standard image benchmarks.

To address this problem, we propose \textbf{SafeECGMatch}, a calibration-aware safe semi-supervised learning framework for single-label ECG classification under label distribution mismatch. Rather than directly adapting image-based Safe SSL to physiological signals, our method is designed around the structural characteristics of ECG time series and the reliability requirements of clinical prediction. SafeECGMatch jointly learns from time- and frequency-domain views of the same ECG sample, applying ECG-specific augmentations. It simultaneously calibrates the predictive confidence in both views, combining the calibrated time- and frequency-domain objectives to enable safer pseudo-label assignment and more reliable OOD rejection.

We make the primary contributions as follows: 1) We introduce the novel Safe SSL framework for ECG classification that explicitly handles label distribution mismatch, to our knowledge; 2) We propose a dual-view calibration mechanism that jointly calibrates signal in both time and frequency domains by integrating them into a single objective function; 3) The proposed approach achieves state-of-the-art performance in both accuracy and calibration, suggesting robust domain adaptation in ECG classification.



\section{Related Works}

\subsection{ECG Classification and Semi-Supervised Learning}
Recent deep learning studies have substantially improved ECG classification in supervised settings. Medformer employs an attention-based Transformer with multi-granularity patching to capture ECG patterns across different temporal scales~\cite{wang2024medformer}. SuPreME couples ECG signals with large-scale physician-written reports and LLM-driven supervision to enable strong multimodal diagnosis~\cite{cai-etal-2025-supreme}; however, this advantage relies on abundant text annotations that are often not available. Trustworthy ECG diagnosis based on OOD detection uses energy scores and ReAct-style activation truncation to filter anomalous inputs~\cite{yu2025trustworthy}, yet this remains a supervised test-time safeguard that rejects unknown samples without exploiting unlabeled data to improve representation learning. 

Semi-supervised learning has also been explored to alleviate the scarcity of expert-labeled recordings. ECGMatch demonstrates that SSL can improve cardiovascular disease prediction~\cite{zhou2023semi}; however, its emphasis on multi-label prediction is not fully aligned with practical single-label settings, which are common in benchmark data (e.g., single-label records constitute the vast majority of PhysioNet/CinC
Challenge benchmarks as evaluated in ECGMatch). Other frameworks tackle label mismatch from an anomaly-detection angle. AnoGAN learns only from normal ECGs and detects deviations as anomalies~\cite{shin2020decision}, while ECG-ADGAN introduces a semi-supervised one-class framework with sequence modeling~\cite{qin2023novel}. These approaches reduce the task to heuristic normal-versus-abnormal detection rather than clinically useful multi-class diagnosis, leaving safe SSL for single-label ECG classification largely unexplored.

\subsection{Safe Semi-Supervised Learning and Calibration}
Because safe semi-supervised ECG classification is largely absent in the literature, the closest precedents come from general safe SSL. OpenMatch extends FixMatch with one-vs-all detector, providing a threshold to detect
outliers with consistency regularization~\cite{sohn2020fixmatch, saito2021openmatch}. IOMatch unifies inlier and outlier utilization by combining a closed-set and multi-binary classifier to optimize open-set targets jointly~\cite{li2023iomatch}. SCOMatch mitigates data overtrusting by scoring unlabeled examples via MSP, maintaining an OOD memory queue for $(K+1)$-class self-training~\cite{wang2024scomatch}. SafeStudent follows a teacher-student route, using energy-discrepancy for safer teacher predictions and label distribution learning to reduce student overconfidence~\cite{he2022safe}. Adello explicitly targets proper calibration through flexible distribution alignment, distilling soft pseudo-labels from underconfident samples to manage severe distribution shifts~\cite{sanchez2024flexible}. 

However, these methods often suffer from overconfidence, which can lead to the generation of incorrect pseudo-labels. CaliMatch addresses this issue by incorporating calibration directly into safe SSL using learnable temperature scaling and adaptive label smoothing derived from validation reliability statistics~\cite{bae2025calimatch}. While highly effective in the image domain, this formulation does not exploit the temporal and spectral structures central to ECG data. SafeECGMatch builds on this calibration-aware formulation but extends it to joint time-domain and frequency-domain learning.

\subsection{Temporal-Frequency Co-Training}
Temporal-frequency co-training for time series semi-supervised learning (TS-TFC) is the closest prior work to our dual-view motivation~\cite{liu2023temporal}. It treats time-domain and frequency-domain signals as complementary views, learns separate encoders through label propagation supported by a temporal-frequency supervised contrastive objective. More recent approaches such as CompleMatch further improved temporal-frequency SSL via complementary co-training and pseudo-label matching \cite{liu2025complematch}. However, these methods do not address label distribution mismatch. SafeECGMatch adopts this dual-view intuition but comprehensively extends it to safe ECG SSL with explicit OOD rejection and calibration.

\begin{figure*}[t] 
  \centering
  \includegraphics[page=4, trim=0 320 0 10, clip, width=0.9\textwidth]{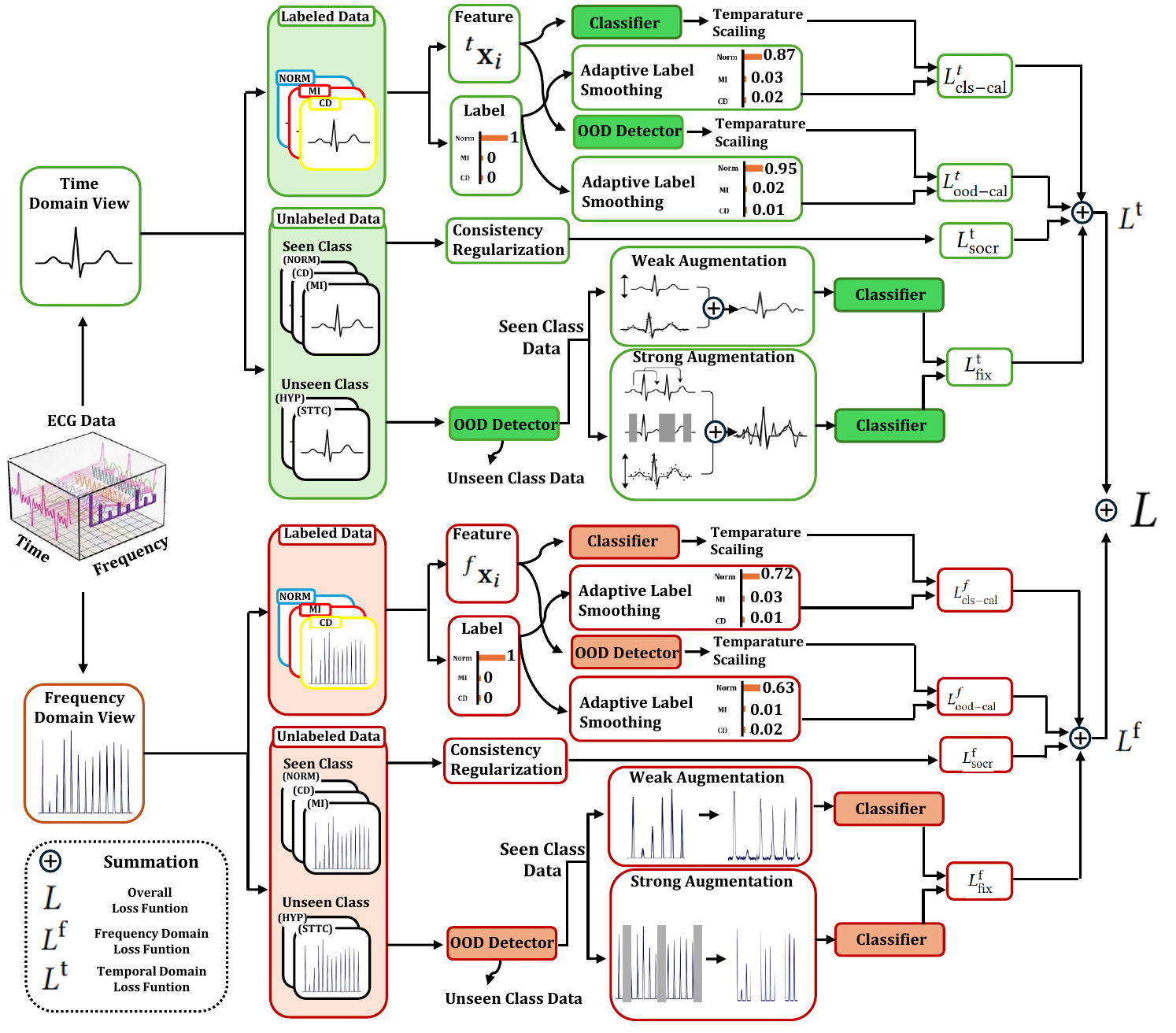}
  \caption{SafeECGMatch Flowchart}
  \label{fig:framework}
\end{figure*}

\section{Methods}

This section presents \textbf{SafeECGMatch}, our calibration-aware safe semi-supervised learning framework for single-label ECG classification under label distribution mismatch. The method is designed for the practical setting in which the unlabeled pool contains both in-distribution ECGs from the target label space and out-of-distribution recordings that should not be converted into pseudo-label supervision. SafeECGMatch introduces a dual-domain design that jointly optimizes a time-domain branch and a frequency-domain branch, together with ECG-aware augmentations and simultaneous calibration in both domains.

\subsection{Problem Setting}

Given a set of ECG 
$^{tem}\mathcal{X} = \{ ^{t}\mathbf{x}_i \}_{i=1}^N
$, and we denote $\mathcal{D}_L=\{(^{t}\mathbf{x}^{\ell}_i, y_i)\}_{i=1}^{N_L}$ as a labeled ECG set, where $^{t}\mathbf{x}^{\ell}_i \in \mathbb{R}^{C \times T}$ is a $C$-lead segment of length $T$ and $y_i \in \{1,\dots,K\}$ is its single class label. Let $\mathcal{D}_U=\{^{t}\mathbf{x}^{u}_j\}_{j=1}^{N_U}$ denote an unlabeled set whose hidden labels may either belong to the known target set or fall outside it, where $N_L + N_U = N$. This is the safe SSL setting with label distribution mismatch.

SafeECGMatch maintains two parallel predictors, one for the time domain(t) and one for the frequency domain(f). For branch $d \in \{\mathrm{t},\mathrm{f}\}$, let $h_{d}(\cdot)$ be a feature encoder, $c_{d}(\cdot)$ be a multiclass classifier, and $r_{d}(\cdot)$ be a OOD detector. The frequency input is obtained by applying a magnitude transformation using the real Fast Fourier Transform(FFT) to the original signal, i.e.,
\begin{equation}
^{frq}\mathcal{X} = \left|\operatorname{rFFT}(\mathbf{^{tem}\mathcal{X}}) \right| = \{ ^{f}\mathbf{x}_i \}_{i=1}^N.
\label{eq:fft}
\end{equation}
For any input ${}^{d}\mathbf{x}$ in branch $d$, where k is the class of labeled ECG set, the posterior of multiclass classifier is defined as
\begin{equation}
p_{d,k}(^{d}\mathbf{x}_i) = \left[\operatorname{softmax}(c_d(h_d(^{d}\mathbf{x}_i)))\right]_k.
\label{eq:posterior}
\end{equation}
OOD detector produces probabilities to distinguish OOD:
\begin{equation}
 q^{}_{d,k}(^{d}\mathbf{x}_i\bigr)
= \operatorname{softmax}\bigl(r_{d,k}(h_d(^{d}\mathbf{x}_i))\bigr).
\label{eq:ovr}
\end{equation}

\subsection{ECG-Specific Dual-Domain Augmentation}
Applying weak and strong augmentations to unlabeled data is crucial for enhancing classifier consistency. Therefore, we propose a specialized augmentation strategy tailored for this purpose.

The weak augmentation for time branch, is generated by per-lead amplitude scaling followed by additive Gaussian noise:
\begin{equation}
\mathcal{A}^{\mathrm{t}}_{w}(^{t}\mathbf{x}_i) = \mathbf{a} \odot (^{t}\mathbf{x}_i) + \boldsymbol{\epsilon},
\quad
\mathbf{a}\sim \mathcal{N}(1, 0.1^2),
\quad
\boldsymbol{\epsilon} \sim \mathcal{N}(0, 0.01^2),
\label{eq:weak-time-aug}
\end{equation}
where $\mathbf{a}\in\mathbb{R}^{C\times 1}$ is broadcast along time, $\boldsymbol{\epsilon}\in\mathbb{R}^{C\times T}$, and $\odot$ denotes channel-wise multiplication. This augmentation perturbs amplitude mildly while preserving overall rhythm and morphology.

The strong augmentation for time branch, first applies harder structural corruption and then uses the weak augmentation. Let $\mathcal{P}_{5}(\cdot)$ denote random permutation of five contiguous temporal segments, and let $\mathcal{M}^{\mathrm{t}}_{s,\ell}(\cdot)$ denote zero-masking on interval $[s,s+\ell)$ with $\ell \sim \mathrm{Unif}\{50,\ldots,200\}$. With independent Bernoulli variables $b_{p},b_{m}\sim\mathrm{Bernoulli}(0.5)$,
\begin{equation}
\mathcal{A}^{\mathrm{t}}_{s}(^{t}\mathbf{x}_i) = \mathcal{A}^{\mathrm{t}}_{w}
\Bigl(
\bigl(\mathcal{M}^{\mathrm{t}}_{s,\ell}\bigr)^{b_m}
\bigl(\mathcal{P}_{5}\bigr)^{b_p}
(^{t}\mathbf{x}_i)
\Bigr),
\label{eq:strong-time-aug}
\end{equation}
where exponent $0$ means the identity map and exponent $1$ means the operation is applied. Thus the strong augmentation used for consistency learning may include segment permutation, temporal masking, or both, followed by the same amplitude-noise perturbation in Eq.~\eqref{eq:weak-time-aug}.

The weak augmentation of frequency branch inherits the weak augmentation from the time branch and maps them into the spectral space using Eq.~\eqref{eq:fft}. Specifically,
\begin{equation}\mathcal{A}^{\mathrm{f}}_{w}(^{f}\mathbf{x}_i) = \left|\operatorname{rFFT}(\mathbf{\mathcal{A}^{\mathrm{t}}_{w}(^{t}\mathbf{x}_i)})\right|
\end{equation}
The strong augmentation for the frequency branch, $\mathcal{A}^{\mathrm{f}}_{s}$, is defined by masking a contiguous frequency band in the magnitude spectrum. Let $L_f$ denote the number of frequency bins, $m=\lfloor 0.15 L_f \rfloor$, and let $s_f$ be sampled uniformly from $\{0,\ldots,L_f-m-1\}$. Then, for each frequency-bin index $j \in \{0,\ldots,L_f-1\}$,
\begin{equation}
\bigl[\mathcal{A}^{\mathrm{f}}_{s}(\mathbf{x}_i^{f})\bigr]_j=
\begin{cases}
0, & s_f \le j < s_f + m,\\
\bigl[\mathcal{A}^{\mathrm{f}}_{w}(\mathbf{x}_i^{f})\bigr]_j, & \text{otherwise}.
\end{cases}
\label{eq:freq-mask-def}
\end{equation} 

\subsection{Calibration-Aware Safe SSL in Each Branch}

The same training routine is applied independently to the time branch and the frequency branch. For branch $d \in \{\mathrm{t},\mathrm{f}\}$, labeled data minibatch $^{d}\mathcal{B}_{\ell}$ and unlabeled data minibatch $^{d}\mathcal{B}_{u}$, the supervised classification loss is
\begin{equation}
L_{\mathrm{cls}}^{d} = -
\sum_{i=1}^{|^{d}\mathcal{B}_{\ell}|}\sum_{k=1}^{K}{y}_{ik} \log p_{d,k}(^{d}\mathbf{x}_{i}^{l}).
\label{eq:lsup}
\end{equation}
The OOD detector is trained on labeled data using
\begin{equation}
\begin{aligned}
L_{\mathrm{ood}}^{d} =- 
\sum_{i=1}^{|^{d}\mathcal{B}_{\ell}|}[\sum_{k=1}^{K}({y}_{ik}
\log q_{d,k}(^{d}\mathbf{x}_{i}^{l})
)+ \min_{l \neq k}\log(1- q_{d,l}(^{d}\mathbf{x}_{i}^{l})\bigr)].
\end{aligned}
\label{eq:lovr}
\end{equation}
For unlabeled data, OOD detector is regularized by soft consistency regularization about the two weak views:
\begin{equation}
\begin{aligned}
L_{\mathrm{socr}}^{d} = \sum_{i=1}^{|^{d}\mathcal{B}_{u}|}\sum_{k=1}^{K}
\|\mathbf{q}_{d,k}(\mathcal{A}^{\mathrm{d}}_{w(1)}(^{d}\mathbf{x}_i^{u})) - \mathbf{q}_{d,k}(\mathcal{A}^{\mathrm{d}}_{w(2)}(^{d}\mathbf{x}_i^{u}))\|_2^2.
\end{aligned}
\label{eq:lsocr}
\end{equation}

After warm-up, SafeECGMatch activates calibration losses for both the multiclass logits and the OOD logits in each branch. Let $T^{d}_{\mathrm{cls}}>0$ and $T^{d}_{\mathrm{ood}}>0$ be learnable temperature parameters. The calibrated classifier and OOD probabilities are
\begin{equation}
\bar{p}_{d,k}(^{d}\mathbf{x}_{i})
= \left[\operatorname{softmax}\left(c_d(h_d(^{d}\mathbf{x}_{i}))/T^{d}_{\mathrm{cls}}\right)\right]_k,
\label{eq:scaled-cls}
\end{equation}
\begin{equation}
\bar{q}_{d,k}(^{d}\mathbf{x}_{i})
= \operatorname{softmax}\left(r_{d,k}(h_d(^{d}\mathbf{x}_{i}))/T^{d}_{\mathrm{ood}}\right).
\label{eq:scaled-ovr}
\end{equation}
Following \cite{bae2025calimatch}, we estimate confidence-conditioned accuracy on the validation split separately for the classifier and the OOD detector to define adaptive label smoothing. The resulting branch-specific target confidences are denoted by $\alpha^{d}_{i}$ for the multiclass head and $\beta^{d}_{i}$ for the OOD detector. Then the calibration loss functions become
\begin{equation}
L_{\mathrm{cls-cal}}^{d} = - \sum_{i=1}^{|^{d}\mathcal{B}_{\ell}|} \sum_{k=1}^{K} \left( \alpha^{d}_{i} \cdot y_{ik} + \frac{1-\alpha^{d}_{i}}{K-1} (1-y_{ik}) \right) \cdot \log \bar{p}_{d,k}(^{d}\mathbf{x}^{\ell}_{i})
\label{eq:lcal}
\end{equation}

\begin{equation}
\begin{aligned}
L_{\mathrm{ood-cal}}^{d} &= - \sum_{i=1}^{|^{d}\mathcal{B}_{\ell}|} \sum_{k=1}^{K} \Bigg[ \\
&\quad \left( \beta^{d}_{i} y_{ik} + (1-\beta^{d}_{i})(1-y_{ik}) \right) \log \bar{q}_{d,k}(^{d}\mathbf{x}^{\ell}_{i}) \\
&\quad + \min_{k} \bigg( \left( (1-\beta^{d}_{i}) y_{ik} + \beta^{d}_{i} (1-y_{ik}) \right) \\
&\qquad \times \log (1 - \bar{q}_{d,k}(^{d}\mathbf{x}^{\ell}_{i})) \bigg) \Bigg]
\end{aligned}
\label{eq:lovr-cal}
\end{equation}

\subsection{Reliable Unlabeled Selection and Dual-Domain Final Objective}
We now aim to train the model using the reliable unlabeled data after filtering out OOD samples by OOD detector. To this end, separate thresholds are assigned to each branch, and filtering is performed independently for each branch. The procedure for obtaining reliable unlabeled data for each branch is defined as follows:
\begin{equation}
\begin{aligned}
^{d}S^{u}_{i} &= \sum_{k} \bar{p}_{d,k}(^{d}\mathbf{x}^{u}_{i}) \cdot \bar{q}_{d,k}(^{d}\mathbf{x}^{u}_{i}) \\
^{d}C^{u}_{i} &= \max_{k} \bar{p}_{d,k}(^{d}\mathbf{x}^{u}_{i})
\end{aligned}
\end{equation}

\begin{equation}
^{d}\mathcal{B}^{\mathcal{I}}_{u} = \left\{ ^{d}\mathbf{x}^{u}_{i} \in ^{d}\mathcal{B}_{u} : (\ ^{d}S^{u}_{i} > T_1) \land (\ ^{d}C^{u}_{i} > T_2) \right\}
\end{equation}

Following FixMatch, applying weak and strong augmentations to unlabeled data can improve classifier performance \cite{sohn2020fixmatch}. Therefore, we apply the ECG-specific augmentation strategies described in Section 3.2 and compute the following FixMatch loss function:
\begin{equation}
\begin{aligned}
L_{\mathrm{fix}}^{d} &=  \sum_{i=1}^{|^{d}\mathcal{B}^{\mathcal{I}}_{u}|} \sum_{k=1}^{K} \mathbb{I} \left( \operatorname{argmax}_{\ell} p_{d,\ell}(\mathcal{A}^{d}_{w}(^{d}\mathbf{x}^{u}_{i})) = k \right) \\
&\quad \times \log p_{d,k}(\mathcal{A}^{d}_{s}(^{d}\mathbf{x}^{u}_{i}))
\end{aligned}
\end{equation}
The complete branch loss is therefore
\begin{equation}
\begin{aligned}
L^{d} &= L_{\mathrm{cls}}^{d} + \lambda_{\mathrm{ood}}L_{\mathrm{ood}}^{d} + \lambda_{\mathrm{socr}}L_{\mathrm{socr}}^{d} \\
&\quad + \lambda_{\mathrm{cls-cal}}L_{\mathrm{cls-cal}}^{d} + \lambda_{\mathrm{ood-cal}}L_{\mathrm{ood-cal}}^{d} + \lambda_{\mathrm{fix}}L_{\mathrm{fix}}^{d}.
\end{aligned}
\label{eq:branch-loss}
\end{equation}

Unlike other formulations that operate in a single domain, SafeECGMatch computes Eq.~\eqref{eq:branch-loss} for both the time branch and the frequency branch within every iteration. The final loss function is the weighted sum of the two branch losses,
\begin{equation}
L = L^{\mathrm{t}} + \lambda_{\mathrm{sum}}L^{\mathrm{f}}.
\label{eq:total-loss}
\end{equation}
This is the key design difference of SafeECGMatch: time-domain calibration and frequency-domain calibration are both optimized simultaneously, and the final training signal is obtained by adding the two calibrated safe SSL objectives. All hyperparameters, including threshold values ($T_1$, $T_2$) and balancing coefficients ($\lambda_{\mathrm{ood}}$, $\lambda_{\mathrm{socr}}$, $\lambda_{\mathrm{cls\text{-}cal}}$, $\lambda_{\mathrm{ood\text{-}cal}}$, $\lambda_{\mathrm{fix}}$), were optimized for ECG time-series data. Sensitivity analysis of $\lambda_{\mathrm{sum}}$ is provided in Appendix B (Table 4).

\begin{algorithm}[t]
\caption{SafeECGMatch}
\label{alg:safeecgmatch}
\begin{algorithmic}[1]
\STATE \textbf{Input:} $\mathcal{D}_L$ (labeled ECG set), $\mathcal{D}_U$ (unlabeled ECG set), $\mathcal{V}$ (validation set)
\STATE \textbf{Input:} Model parameters $\Theta$ for $\{h_d, c_d, r_d\}_{d\in\{\mathrm{t},\mathrm{f}\}}$
\STATE \textbf{Input:} Hyperparameters $E_{\max}, I_{\max}, E_{\mathrm{warm}}, \eta, T_1, T_2$
\STATE \textbf{Input:} Loss weights $\lambda_{\mathrm{ood}}, \lambda_{\mathrm{socr}}, \lambda_{\mathrm{cls-cal}}, \lambda_{\mathrm{ood-cal}}, \lambda_{\mathrm{fix}}, \lambda_{\mathrm{sum}}$
\FOR{epoch $= 1$ \TO $E_{\max}$}
    \FOR{iteration $= 1$ \TO $I_{\max}$}
        \STATE Sample minibatch $\mathcal{B}_{\ell} \subset \mathcal{D}_L$ and minibatch $\mathcal{B}_{u} \subset \mathcal{D}_U$
        
        \STATE Generate $\mathcal{A}^{\mathrm{t}}_{w(1)}, \mathcal{A}^{\mathrm{t}}_{w(2)}, \mathcal{A}^{\mathrm{t}}_{s}$ from $\mathcal{B}_{u}$ via Eqs.~\eqref{eq:weak-time-aug}, \eqref{eq:strong-time-aug}
        \STATE Generate $\mathcal{A}^{\mathrm{f}}_{w(1)}, \mathcal{A}^{\mathrm{f}}_{w(2)}, \mathcal{A}^{\mathrm{f}}_{s}$ from $\mathcal{B}_{u}$ via Eqs.~(6), \eqref{eq:freq-mask-def}
        
        \FOR{each branch $d \in \{\mathrm{t},\mathrm{f}\}$}
            \STATE Compute $p_{d}$ and $q_{d}$ via Eqs.~\eqref{eq:posterior}, \eqref{eq:ovr}
            \STATE $L_{\mathrm{base}}^{d} \leftarrow L_{\mathrm{cls}}^{d} + \lambda_{\mathrm{ood}}L_{\mathrm{ood}}^{d} + \lambda_{\mathrm{socr}}L_{\mathrm{socr}}^{d}$ \quad (Eqs.~\eqref{eq:lsup}--\eqref{eq:lsocr})
        \ENDFOR
        \IF{epoch $\geq E_{\mathrm{warm}}$}
            \FOR{each branch $d \in \{\mathrm{t},\mathrm{f}\}$}
                \STATE Get calibrated prob. $\bar{p}_{d}, \bar{q}_{d}$ via Eqs.~\eqref{eq:scaled-cls}, \eqref{eq:scaled-ovr}
                \STATE Filter reliable subset ${}^{d}\mathcal{B}^{\mathcal{I}}_{u}$ via Eqs.~(15)--(16)
                \STATE $L_{\mathrm{warm}}^{d} \leftarrow \lambda_{\mathrm{cls-cal}}L_{\mathrm{cls-cal}}^{d} + \lambda_{\mathrm{ood-cal}}L_{\mathrm{ood-cal}}^{d} + \lambda_{\mathrm{fix}}L_{\mathrm{fix}}^{d}$
            \ENDFOR
        \ELSE
            \FOR{each branch $d \in \{\mathrm{t},\mathrm{f}\}$}
                \STATE $L_{\mathrm{warm}}^{d} \leftarrow 0$
            \ENDFOR
        \ENDIF
        
        \FOR{each branch $d \in \{\mathrm{t},\mathrm{f}\}$}
            \STATE $L^{d} \leftarrow L_{\mathrm{base}}^{d} + L_{\mathrm{warm}}^{d}$ \quad (Eq.~\eqref{eq:branch-loss})
        \ENDFOR
        \STATE $L \leftarrow L^{\mathrm{t}} + \lambda_{\mathrm{sum}}L^{\mathrm{f}}$ \quad (Eq.~\eqref{eq:total-loss})
        \STATE Update model parameters with stochastic gradient descent: $\Theta \leftarrow \Theta - \eta \nabla L$
    \ENDFOR
    
    \STATE Update branch-wise statistic ($\alpha^{d}, \beta^{d}$) and temperatures ($T^d_{\mathrm{cls}}, T^d_{\mathrm{ood}}$) via $\mathcal{V}$
\ENDFOR
\RETURN Trained SafeECGMatch model $\{h_d, c_d, r_d\}_{d\in\{\mathrm{t},\mathrm{f}\}}$
\end{algorithmic}
\end{algorithm}

\section{Experiments}

\subsection{Experimental Setup}
We evaluate SafeECGMatch on two 500 Hz single-label ECG benchmarks, PTB-XL~\cite{wagner2020ptb} and CinC21~\cite{reyna2021will}, under open-set semi-supervised learning with label distribution mismatch. 

Following the standard evaluation protocol of~\cite{oliver2018realistic}, all methods utilize a ResNet1D backbone. Models are optimized using Adam with a learning rate of 0.001, a total batch size of 64, and a labeled-to-unlabeled ratio of 1:99. Training spans 50,000 iterations with a 500-iteration warm-up, and checkpoints are evaluated every 1,000 iterations. Each experiment is repeated with three random seeds, and we report the mean and standard deviation. The evaluation metrics include classification accuracy (ACC) alongside calibration metrics: Expected Calibration Error (ECE), Adaptive Calibration Error (ACE), and Static Calibration Error (SCE).

For both benchmarks, we retain only single-label ECGs to construct five-class datasets, applying a random 8:1:1 train/validation/test split, with unseen class data of unlabeled data proportion set to 30\% or 60\%. 

\begin{itemize}
    \item \textbf{PTB-XL:} PTB-XL \cite{wagner2020ptb} is an open-access dataset of clinical 12-lead ECG records annotated by cardiologists. For this dataset, we define Normal ECG (NORM), Myocardial Infarction (MI), and Conduction Disturbance (CD) as seen classes, while ST/T-Change (STTC) and Hypertrophy (HYP) are treated as unseen classes. To retain the unlabeled data proportion, we utilized a subset of 15,000 ECGs. This dataset is partitioned into 150 labeled training samples (50 per seen class) and an unlabeled pool of 14,850 ECGs.
    
    \item \textbf{CinC21:} The PhysioNet/Computing in Cardiology (CinC) Challenge 2021 dataset is a comprehensive multi-center corpus used to evaluate automated cardiac abnormality identification~\cite{reyna2021will}. We derive a single-label subset from this collection, preprocessed via the ECGMatch pipeline~\cite{zhou2023semi}. Assigning NORM, Abnormal Rhythm (RHY), and CD as seen classes, and ST/T Abnormality (ST) and Other Abnormalities (OTHER) as unseen classes results in 31,138 total ECGs. This comprises 312 labeled samples (104 per seen class) and 30,826 unlabeled samples.
\end{itemize}

We evaluate SafeECGMatch together with the supervised and latest competing methods--FixMatch~\cite{sohn2020fixmatch}, IOMatch~\cite{li2023iomatch}, OpenMatch~\cite{saito2021openmatch}, SCOMatch~\cite{wang2024scomatch}, SafeStudent~\cite{he2022safe}, ECGMatch~\cite{zhou2023semi}, Adello~\cite{sanchez2024flexible}, CaliMatch~\cite{bae2025calimatch}, TS-TFC \cite{liu2023temporal}, CompleMatch \cite{liu2025complematch}. Since safe ECG learning requires both correct classification and reliable confidence under label distribution mismatch~\cite{liu2022devil}, we evaluate accuracy and calibration jointly.

\subsection{Benchmark Results}

Appendix A (Table~\ref{tab:benchmark-results}) summarizes the main benchmark results. SafeECGMatch achieves the best or second-best ACC in all four protocols while also obtaining the best or second-best calibration across all settings. On PTB-XL with 60\% OOD, it improves ACC from 0.8360 to 0.8433 and reduces ECE from 0.054 to 0.047. On CinC21 with 60\% OOD, it improves ACC from 0.7617 to 0.7803 and simultaneously lowers ECE from 0.1030 to 0.0860. Crucially, by mitigating the overconfidence typically seen when predicting unseen classes, our model successfully filters harmful out-of-distribution signals to prevent performance degradation. These results indicate that the proposed dual-domain design fundamentally improves the accuracy-calibration trade-off rather than merely trading one for the other.

\begin{table}[t]
    \centering
    \caption{Summary of benchmark results on PTB-XL~\cite{wagner2020ptb} and CinC21~\cite{reyna2021will}. Best results are in bold, and second-best results are underlined. See full benchmark with additional methods and metrics (SCE, ACE) in Appendix A.}
    \label{tab:benchmark-summary}
    
    \resizebox{\columnwidth}{!}{%
    \setlength{\tabcolsep}{1.5pt}
    \renewcommand{\arraystretch}{1.1}
    
    \begin{tabular}{@{}llcccc@{}}
        \toprule
        & & \multicolumn{2}{c}{\textbf{PTB-XL~\cite{wagner2020ptb}}} & \multicolumn{2}{c}{\textbf{CinC21~\cite{reyna2021will}}} \\
        \cmidrule(lr){3-4} \cmidrule(lr){5-6}
        \textbf{Method} & \textbf{OOD} & \textbf{ACC} & \textbf{ECE} & \textbf{ACC} & \textbf{ECE} \\
        \midrule
        
        \multirow{2}{*}{Supervised} 
         & 30\% & \multirow{2}{*}{0.734$\pm$0.040} & \multirow{2}{*}{0.236$\pm$0.041} & \multirow{2}{*}{0.687$\pm$0.053} & \multirow{2}{*}{0.249$\pm$0.029} \\
         & 60\% & & & & \\

        \multirow{2}{*}{OpenMatch~\cite{saito2021openmatch}} 
         & 30\% & 0.831$\pm$0.007 & 0.158$\pm$0.008 & 0.779$\pm$0.050 & 0.202$\pm$0.043 \\
         & 60\% & \underline{0.830$\pm$0.004} & 0.155$\pm$0.005 & 0.749$\pm$0.050 & 0.218$\pm$0.030 \\

        \multirow{2}{*}{CompleMatch~\cite{liu2025complematch}} 
         & 30\% & 0.750$\pm$0.017 & 0.160$\pm$0.014 & \underline{0.795$\pm$0.006} & 0.135$\pm$0.006 \\
         & 60\% & 0.734$\pm$0.032 & 0.183$\pm$0.018 & \textbf{0.786$\pm$0.011} & 0.124$\pm$0.038 \\

        \multirow{2}{*}{ECGMatch~\cite{zhou2023semi}} 
         & 30\% & 0.788$\pm$0.039 & 0.072$\pm$0.011 & 0.635$\pm$0.049 & 0.092$\pm$0.014 \\
         & 60\% & 0.791$\pm$0.033 & 0.122$\pm$0.039 & 0.651$\pm$0.185 & 0.132$\pm$0.097 \\

        \multirow{2}{*}{CaliMatch~\cite{bae2025calimatch}} 
         & 30\% & \underline{0.845$\pm$0.002} & \textbf{0.039$\pm$0.010} & \underline{0.819$\pm$0.010} & \underline{0.057$\pm$0.005} \\
         & 60\% & 0.831$\pm$0.004 & \underline{0.054$\pm$0.004} & 0.758$\pm$0.006 & \underline{0.103$\pm$0.021} \\

        \multirow{2}{*}{\textbf{SafeECGMatch}} 
         & 30\% & \textbf{0.852$\pm$0.005} & \underline{0.045$\pm$0.015} & \textbf{0.829$\pm$0.009} & \textbf{0.052$\pm$0.014} \\
         & 60\% & \textbf{0.843$\pm$0.009} & \textbf{0.047$\pm$0.011} & \underline{0.780$\pm$0.018} & \textbf{0.086$\pm$0.027} \\

        \bottomrule
    \end{tabular}%
    }
\end{table}

\begin{figure}[H] 
  \centering
  \includegraphics[page=5, trim=0 550 0 20, clip, width=\linewidth]{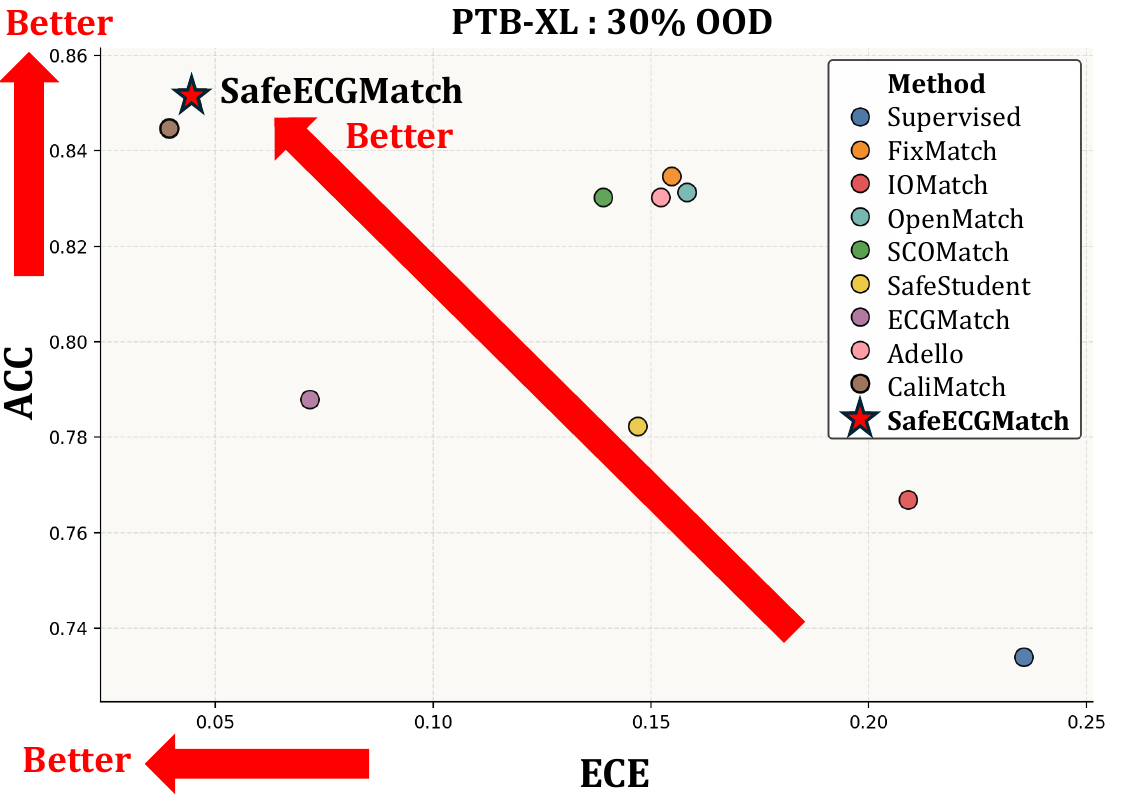}
  \caption{Accuracy-calibration trade-off on PTB-XL with 30\% OOD contamination. Since calibration-aware learning often involves an ACC-ECE trade-off, the preferred model should balance discriminative performance and confidence reliability rather than prioritizing either metric in isolation~\cite{wang2022calibrating, liu2022devil}. SafeECGMatch provides a better compromise by achieving the highest ACC with only a slight increase in ECE.}
  \label{fig:framework}
\end{figure}

\subsection{Ablation Study}
We conducted an ablation study under the severe 60\% OOD setting. We compare: (1) \textbf{SafeECGMatch}; (2) \textbf{w/o frequency-domain calibration} (analogous to CaliMatch); and (3) \textbf{w/o frequency and time-domain calibration} (analogous to OpenMatch). As shown in Table~\ref{tab:ablation}, explicitly imposing calibration constraints on both domains significantly outperforms merely adding an uncalibrated auxiliary spectral branch. This confirms that joint time and frequency calibration is essential for simultaneously maximizing discriminative performance and confidence reliability.

\begin{table}[htbp]
    \centering
    \caption{Ablation study at 60\% OOD on PTB-XL \& CinC21}
    \label{tab:ablation}
    
    \resizebox{\columnwidth}{!}{
    \setlength{\tabcolsep}{2pt} 
    \renewcommand{\arraystretch}{1.2} 
    
    \begin{tabular}{@{}lcccc@{}}
        \toprule
        \multirow{2}{*}{\textbf{Method}} & \multicolumn{2}{c}{\textbf{PTB-XL}} & \multicolumn{2}{c}{\textbf{CinC21}} \\
        \cmidrule(lr){2-3} \cmidrule(lr){4-5}
        & \textbf{ACC} & \textbf{ECE} & \textbf{ACC} & \textbf{ECE} \\
        \midrule
        \textbf{SafeECGMatch} & \best{0.843$\pm$0.007} & \best{0.047$\pm$0.009} & \best{0.780$\pm$0.014} & \best{0.086$\pm$0.022} \\
        w/o Freq. Calib. & \second{0.831$\pm$0.003} & \second{0.054$\pm$0.004} & \second{0.758$\pm$0.005} & \second{0.103$\pm$0.017} \\
        w/o both Calib. & 0.830$\pm$0.004 & 0.155$\pm$0.004 & 0.749$\pm$0.041 & 0.218$\pm$0.024 \\
        \bottomrule
    \end{tabular}%
    }
\end{table}

\section{Conclusion}
We investigated approaches to safely and reliably train models for ECG data to predict cardiovascular disorders—such as arrhythmias, myocardial infarction, and heart failure—in more realistic scenarios. Many cardiovascular conditions, such as heart failure, do not exhibit symptoms at rest but become apparent during physical activity. Consequently, diagnostic procedures often involve a treadmill test to measure ECG, heart rate, and blood pressure during exertion. In such dynamic settings, noise frequently occurs, and OOD instances—unseen classes presenting completely new morphologies—inevitably emerge. When training models with such contaminated data, there is a critical risk of misclassifying actual diseases as normal due to overconfident, erroneous predictions.

SafeECGMatch directly addresses this issue by reliably filtering out unseen classes and reducing model overconfidence through explicit calibration, yielding more accurate predictions for known ID diagnostic groups (e.g., NORM, CD, MI). Across PTB-XL and CinC21, our approach achieves state-of-the-art accuracy and calibration under label distribution mismatch, while using only 1\% of labeled data—substantially reducing the annotation costs required from clinical experts.

More broadly, since many time-series problems admit multiple informative views (e.g., time-frequency, sensor-wise), SafeECGMatch can be seen as one instance of a calibration-aware multi-view safe SSL framework, suggesting its applicability to wider time-series settings with abundant unlabeled data and hidden distribution shift.  

At the same time, this study has several limitations. Our experiments focus on single-label ECG classification, two benchmark datasets, and a fixed time-frequency two-view design. The current framework also does not yet address more realistic settings such as multi-label diagnosis. These limitations define the next step of our work. In future research, we plan to extend SafeECGMatch to multi-label classification, and evaluate the framework on broader medical and general time-series tasks.

\begin{acks}
This work was supported by the National Research Foundation of Korea (NRF) grants funded by the Ministry of Science and ICT (MSIT) (RS-2024-00455720, RS-2026-25481174); the National Institute of Health (NIH) research project (2026-ER0904-00); the Korea Health Technology R\&D Project through the Korea Health Industry Development Institute (KHIDI), funded by the Ministry of Health \& Welfare, Korea (RS-2025-02220534); the Advanced GPU Utilization Support Program and the High-Performance Computing Support project (RQT-25-070083), both funded by MSIT, Korea; and Hankuk University of Foreign Studies Research Fund of 2026.
\end{acks}

\bibliographystyle{ACM-Reference-Format}
\clearpage
\bibliography{ref}

@String{Computing = "Computing" }

@String{Computer = "{IEEE} Computer" }

@String{Springer = "Springer-Verlag" }

@ArtifactSoftware{R,
    title = {R: A Language and Environment for Statistical Computing},
    author = {{R Core Team}},
    organization = {R Foundation for Statistical Computing},
    address = {Vienna, Austria},
    year = {2019},
    url = {https://www.R-project.org/},
}

@article{wang2024medformer,
  title={Medformer: A multi-granularity patching transformer for medical time-series classification},
  author={Wang, Yihe and Huang, Nan and Li, Taida and Yan, Yujun and Zhang, Xiang},
  journal={Advances in Neural Information Processing Systems},
  volume={37},
  pages={36314--36341},
  year={2024}
}

@article{sohn2020fixmatch,
  title={Fixmatch: Simplifying semi-supervised learning with consistency and confidence},
  author={Sohn, Kihyuk and Berthelot, David and Carlini, Nicholas and Zhang, Zizhao and Zhang, Han and Raffel, Colin A and Cubuk, Ekin Dogus and Kurakin, Alexey and Li, Chun-Liang},
  journal={Advances in neural information processing systems},
  volume={33},
  pages={596--608},
  year={2020}
}

@article{saito2021openmatch,
  title={Openmatch: Open-set semi-supervised learning with open-set consistency regularization},
  author={Saito, Kuniaki and Kim, Donghyun and Saenko, Kate},
  journal={Advances in neural information processing systems},
  volume={34},
  pages={25956--25967},
  year={2021}
}

@inproceedings{li2023iomatch,
  title={Iomatch: Simplifying open-set semi-supervised learning with joint inliers and outliers utilization},
  author={Li, Zekun and Qi, Lei and Shi, Yinghuan and Gao, Yang},
  booktitle={Proceedings of the IEEE/CVF international conference on computer vision},
  pages={15870--15879},
  year={2023}
}

@inproceedings{wang2024scomatch,
  title={Scomatch: Alleviating overtrusting in open-set semi-supervised learning},
  author={Wang, Zerun and Xiang, Liuyu and Huang, Lang and Mao, Jiafeng and Xiao, Ling and Yamasaki, Toshihiko},
  booktitle={European Conference on Computer Vision},
  pages={217--233},
  year={2024},
  organization={Springer}
}

@inproceedings{he2022safe,
  title={Safe-student for safe deep semi-supervised learning with unseen-class unlabeled data},
  author={He, Rundong and Han, Zhongyi and Lu, Xiankai and Yin, Yilong},
  booktitle={Proceedings of the IEEE/CVF conference on computer vision and pattern recognition},
  pages={14585--14594},
  year={2022}
}

@inproceedings{sanchez2024flexible,
  title={Flexible distribution alignment: Towards long-tailed semi-supervised learning with proper calibration},
  author={Sanchez Aimar, Emanuel and Helgesen, Nathaniel and Xu, Yonghao and Kuhlmann, Marco and Felsberg, Michael},
  booktitle={European Conference on Computer Vision},
  pages={307--327},
  year={2024},
  organization={Springer}
}

@inproceedings{bae2025calimatch,
  title={CaliMatch: Adaptive Calibration for Improving Safe Semi-supervised Learning},
  author={Bae, Jinsoo and Kim, Seoung Bum and Do, Hyungrok},
  booktitle={Proceedings of the IEEE/CVF International Conference on Computer Vision},
  pages={2867--2876},
  year={2025}
}

@article{zhou2023semi,
  title={Semi-supervised learning for multi-label cardiovascular diseases prediction: a multi-dataset study},
  author={Zhou, Rushuang and Lu, Lei and Liu, Zijun and Xiang, Ting and Liang, Zhen and Clifton, David A and Dong, Yining and Zhang, Yuan-Ting},
  journal={IEEE Transactions on Pattern Analysis and Machine Intelligence},
  volume={46},
  number={5},
  pages={3305--3320},
  year={2023},
  publisher={IEEE}
}

@article{yu2025trustworthy,
  title={Trustworthy diagnosis of Electrocardiography signals based on out-of-distribution detection},
  author={Yu, Bowen and Liu, Yuhong and Wu, Xin and Ren, Jing and Zhao, Zhibin},
  journal={Plos one},
  volume={20},
  number={2},
  pages={e0317900},
  year={2025},
  publisher={Public Library of Science San Francisco, CA USA}
}

@article{shin2020decision,
  title={Decision boundary-based anomaly detection model using improved AnoGAN from ECG data},
  author={Shin, Dong-Hoon and Park, Roy C and Chung, Kyungyong},
  journal={IEEE Access},
  volume={8},
  pages={108664--108674},
  year={2020},
  publisher={IEEE}
}

@article{qin2023novel,
  title={A novel temporal generative adversarial network for electrocardiography anomaly detection},
  author={Qin, Jing and Gao, Fujie and Wang, Zumin and Wong, David C and Zhao, Zhibin and Relton, Samuel D and Fang, Hui},
  journal={Artificial Intelligence in Medicine},
  volume={136},
  pages={102489},
  year={2023},
  publisher={Elsevier}
}

@inproceedings{liu2023temporal,
  title={Temporal-frequency co-training for time series semi-supervised learning},
  author={Liu, Zhen and Ma, Qianli and Ma, Peitian and Wang, Linghao},
  booktitle={Proceedings of the AAAI conference on artificial intelligence},
  volume={37},
  number={7},
  pages={8923--8931},
  year={2023}
}

@article{oliver2018realistic,
  title={Realistic evaluation of deep semi-supervised learning algorithms},
  author={Oliver, Avital and Odena, Augustus and Raffel, Colin A and Cubuk, Ekin Dogus and Goodfellow, Ian},
  journal={Advances in neural information processing systems},
  volume={31},
  year={2018}
}

@article{wagner2020ptb,
  title={PTB-XL, a large publicly available electrocardiography dataset},
  author={Wagner, Patrick and Strodthoff, Nils and Bousseljot, Ralf-Dieter and Kreiseler, Dieter and Lunze, Fatima I and Samek, Wojciech and Schaeffter, Tobias},
  journal={Scientific data},
  volume={7},
  number={1},
  pages={154},
  year={2020},
  publisher={Nature Publishing Group UK London}
}

@inproceedings{reyna2021will,
  title={Will two do? varying dimensions in electrocardiography: The physionet/computing in cardiology challenge 2021},
  author={Reyna, Matthew A and Sadr, Nadi and Alday, Erick A Perez and Gu, Annie and Shah, Amit J and Robichaux, Chad and Rad, Ali Bahrami and Elola, Andoni and Seyedi, Salman and Ansari, Sardar and others},
  booktitle={2021 computing in cardiology (CinC)},
  volume={48},
  pages={1--4},
  year={2021},
  organization={IEEE}
}

@inproceedings{liu2022devil,
  title={The devil is in the margin: Margin-based label smoothing for network calibration},
  author={Liu, Bingyuan and Ben Ayed, Ismail and Galdran, Adrian and Dolz, Jose},
  booktitle={Proceedings of the IEEE/CVF Conference on Computer Vision and Pattern Recognition},
  pages={80--88},
  year={2022}
}

@inproceedings{cai-etal-2025-supreme,
    title = "{S}u{P}re{ME}: A Supervised Pre-training Framework for Multimodal {ECG} Representation Learning",
    author = "Cai, Mingsheng  and
      Jiang, Jiuming  and
      Huang, Wenhao  and
      Liu, Che  and
      Arcucci, Rossella",
    editor = "Christodoulopoulos, Christos  and
      Chakraborty, Tanmoy  and
      Rose, Carolyn  and
      Peng, Violet",
    booktitle = "Findings of the Association for Computational Linguistics: EMNLP 2025",
    month = nov,
    year = "2025",
    address = "Suzhou, China",
    publisher = "Association for Computational Linguistics",
    url = "https://aclanthology.org/2025.findings-emnlp.633/",
    doi = "10.18653/v1/2025.findings-emnlp.633",
    pages = "11817--11844",
    ISBN = "979-8-89176-335-7"
}

@article{attia2019screening,
  title={Screening for cardiac contractile dysfunction using an artificial intelligence--enabled electrocardiogram},
  author={Attia, Zachi I and Kapa, Suraj and Lopez-Jimenez, Francisco and McKie, Paul M and Ladewig, Dorothy J and Satam, Gaurav and Pellikka, Patricia A and Enriquez-Sarano, Maurice and Noseworthy, Peter A and Munger, Thomas M and others},
  journal={Nature medicine},
  volume={25},
  number={1},
  pages={70--74},
  year={2019},
  publisher={Nature Publishing Group US New York}
}

@article{vrints20242024,
  title={2024 ESC guidelines for the management of chronic coronary syndromes: developed by the task force for the management of chronic coronary syndromes of the European Society of Cardiology (ESC) endorsed by the European Association for Cardio-Thoracic Surgery (EACTS)},
  author={Vrints, Christiaan and Andreotti, Felicita and Koskinas, Konstantinos C and Rossello, Xavier and Adamo, Marianna and Ainslie, James and Banning, Adrian Paul and Budaj, Andrzej and Buechel, Ronny R and Chiariello, Giovanni Alfonso and others},
  journal={European heart journal},
  volume={45},
  number={36},
  pages={3415--3537},
  year={2024},
  publisher={Oxford University Press UK}
}

@inproceedings{wang2022calibrating,
  title={Calibrating imbalanced classifiers with focal loss: An empirical study},
  author={Wang, Cheng and Balazs, Jorge and Szarvas, Gy{\"o}rgy and Ernst, Patrick and Poddar, Lahari and Danchenko, Pavel},
  booktitle={Proceedings of the 2022 Conference on Empirical Methods in Natural Language Processing: Industry Track},
  pages={145--153},
  year={2022}
}

@article{kim2026morcu,
  title={MORCU: Margin-based ordinal classification with dynamic regularization for calibration and unimodality},
  author={Kim, Daehwan and Chung, Haejun and Jang, Ikbeom},
  journal={Pattern Recognition},
  pages={113820},
  year={2026},
  publisher={Elsevier}
}

@article{liu2025complematch,
  title={CompleMatch: Boosting Time-Series Semi-Supervised Classification With Temporal-Frequency Complementarity},
  author={Liu, Zhen and Zeng, Kun and Ma, Qianli and Kwok, James T},
  journal={IEEE Transactions on Pattern Analysis and Machine Intelligence},
  year={2025},
  publisher={IEEE}
}

\clearpage
\onecolumn

\appendix

\section{Detailed Benchmark Results}
\begin{table}[H]
    \centering
    \caption{Main benchmark results on PTB-XL~\cite{wagner2020ptb} and CinC21~\cite{reyna2021will}. Best results are in \textbf{bold} and second-best results are underlined for each metric within each OOD setting. Lower ECE, ACE, and SCE represent better calibration.}
    \label{tab:benchmark-results}
    
    \resizebox{\textwidth}{!}{%
    \setlength{\tabcolsep}{5pt}
    \renewcommand{\arraystretch}{1.15} 
    
    \begin{tabular}{llcccccccc}
        \toprule
        \multirow{2}{*}{\textbf{Method}} & \multirow{2}{*}{\textbf{OOD}} & \multicolumn{4}{c}{\textbf{PTB-XL~\cite{wagner2020ptb}}} & \multicolumn{4}{c}{\textbf{CinC21~\cite{reyna2021will}}} \\
        \cmidrule(lr){3-6} \cmidrule(lr){7-10}
        & & \textbf{ACC $\uparrow$} & \textbf{ECE $\downarrow$} & \textbf{ACE $\downarrow$} & \textbf{SCE $\downarrow$} & \textbf{ACC $\uparrow$} & \textbf{ECE $\downarrow$} & \textbf{ACE $\downarrow$} & \textbf{SCE $\downarrow$} \\
        \midrule
        
        \multirow{2}{*}{Supervised} 
         & 30\% & \multirow{2}{*}{0.734$\pm$0.040} & \multirow{2}{*}{0.236$\pm$0.041} & \multirow{2}{*}{0.160$\pm$0.030} & \multirow{2}{*}{0.164$\pm$0.028} & \multirow{2}{*}{0.687$\pm$0.053} & \multirow{2}{*}{0.249$\pm$0.029} & \multirow{2}{*}{0.169$\pm$0.020} & \multirow{2}{*}{0.172$\pm$0.021} \\
         & 60\% & & & & & & & & \\ \addlinespace
        
        FixMatch~\cite{sohn2020fixmatch} 
         & 30\% & 0.835$\pm$0.010 & 0.155$\pm$0.013 & 0.098$\pm$0.006 & 0.105$\pm$0.008 & 0.787$\pm$0.025 & 0.191$\pm$0.025 & 0.125$\pm$0.016 & 0.129$\pm$0.017 \\
        (NeurIPS'20) & 60\% & 0.819$\pm$0.005 & 0.166$\pm$0.003 & 0.104$\pm$0.004 & 0.113$\pm$0.001 & 0.751$\pm$0.018 & 0.210$\pm$0.032 & 0.142$\pm$0.022 & 0.146$\pm$0.021 \\ \addlinespace
        
        IOMatch~\cite{li2023iomatch} 
         & 30\% & 0.767$\pm$0.036 & 0.209$\pm$0.031 & 0.141$\pm$0.023 & 0.144$\pm$0.022 & 0.796$\pm$0.010 & 0.179$\pm$0.004 & 0.119$\pm$0.006 & 0.123$\pm$0.004 \\
        (ICCV'23) & 60\% & 0.771$\pm$0.024 & 0.208$\pm$0.024 & 0.135$\pm$0.014 & 0.143$\pm$0.016 & 0.762$\pm$0.002 & 0.211$\pm$0.002 & 0.143$\pm$0.002 & 0.146$\pm$0.002 \\ \addlinespace
        
        OpenMatch~\cite{saito2021openmatch} 
         & 30\% & 0.831$\pm$0.007 & 0.158$\pm$0.008 & 0.098$\pm$0.008 & 0.107$\pm$0.005 & 0.779$\pm$0.050 & 0.202$\pm$0.043 & 0.133$\pm$0.033 & 0.138$\pm$0.031 \\
        (NeurIPS'21) & 60\% & 0.830$\pm$0.004 & 0.155$\pm$0.005 & 0.097$\pm$0.002 & 0.106$\pm$0.004 & 0.749$\pm$0.050 & 0.218$\pm$0.030 & 0.147$\pm$0.024 & 0.152$\pm$0.023 \\ \addlinespace
        
        SCOMatch~\cite{wang2024scomatch} 
         & 30\% & 0.830$\pm$0.010 & 0.139$\pm$0.013 & 0.089$\pm$0.006 & 0.097$\pm$0.006 & 0.785$\pm$0.010 & 0.175$\pm$0.004 & 0.118$\pm$0.004 & 0.121$\pm$0.004 \\
        (ECCV'24) & 60\% & \underline{0.836$\pm$0.010} & 0.145$\pm$0.008 & 0.091$\pm$0.009 & 0.100$\pm$0.006 & 0.737$\pm$0.014 & 0.214$\pm$0.009 & 0.145$\pm$0.010 & 0.149$\pm$0.009 \\ \addlinespace
        
        SafeStudent~\cite{he2022safe} 
         & 30\% & 0.782$\pm$0.019 & 0.147$\pm$0.021 & 0.105$\pm$0.019 & 0.108$\pm$0.018 & 0.713$\pm$0.051 & 0.181$\pm$0.039 & 0.124$\pm$0.026 & 0.126$\pm$0.027 \\
        (CVPR'22) & 60\% & 0.771$\pm$0.032 & 0.164$\pm$0.020 & 0.115$\pm$0.016 & 0.119$\pm$0.014 & 0.735$\pm$0.003 & 0.169$\pm$0.008 & 0.129$\pm$0.001 & 0.130$\pm$0.003 \\ \addlinespace
        
        ECGMatch~\cite{zhou2023semi} 
         & 30\% & 0.788$\pm$0.039 & 0.072$\pm$0.011 & 0.065$\pm$0.001 & 0.069$\pm$0.001 & 0.635$\pm$0.049 & 0.092$\pm$0.014 & 0.143$\pm$0.030 & 0.144$\pm$0.029 \\
        (TPAMI'23) & 60\% & 0.791$\pm$0.033 & 0.122$\pm$0.039 & 0.098$\pm$0.042 & 0.103$\pm$0.036 & 0.651$\pm$0.185 & 0.132$\pm$0.097 & 0.159$\pm$0.125 & 0.159$\pm$0.125 \\ \addlinespace
        
        Adello~\cite{sanchez2024flexible} 
         & 30\% & 0.830$\pm$0.008 & 0.152$\pm$0.007 & 0.093$\pm$0.005 & 0.104$\pm$0.004 & 0.771$\pm$0.043 & 0.204$\pm$0.038 & 0.135$\pm$0.028 & 0.139$\pm$0.027 \\
        (ECCV'24) & 60\% & 0.818$\pm$0.006 & 0.165$\pm$0.006 & 0.102$\pm$0.002 & 0.112$\pm$0.004 & 0.739$\pm$0.028 & 0.225$\pm$0.025 & 0.152$\pm$0.017 & 0.155$\pm$0.018 \\ \addlinespace

        TS-TFC~\cite{liu2023temporal} 
         & 30\% & 0.736$\pm$0.045 & 0.175$\pm$0.042 & 0.139$\pm$0.025 & 0.139$\pm$0.025 & 0.783$\pm$0.013 & 0.148$\pm$0.011 & 0.103$\pm$0.012 & 0.105$\pm$0.010 \\
        (AAAI'23) & 60\% & 0.759$\pm$0.030 & 0.162$\pm$0.022 & 0.122$\pm$0.016 & 0.123$\pm$0.016 & 0.775$\pm$0.002 & 0.147$\pm$0.004 & 0.109$\pm$0.004 & 0.111$\pm$0.002 \\ \addlinespace
        
        CompleMatch~\cite{liu2025complematch} 
         & 30\% & 0.750$\pm$0.017 & 0.160$\pm$0.014 & 0.128$\pm$0.013 & 0.129$\pm$0.012 & 0.795$\pm$0.006 & 0.135$\pm$0.006 & 0.094$\pm$0.002 & 0.097$\pm$0.001 \\
        (TPAMI'25) & 60\% & 0.734$\pm$0.032 & 0.183$\pm$0.018 & 0.135$\pm$0.015 & 0.137$\pm$0.013 & \textbf{0.786$\pm$0.011} & 0.124$\pm$0.038 & 0.094$\pm$0.014 & 0.096$\pm$0.014 \\ \addlinespace
        
        CaliMatch~\cite{bae2025calimatch} 
         & 30\% & \underline{0.845$\pm$0.002} & \textbf{0.039$\pm$0.010} & \underline{0.043$\pm$0.006} & \underline{0.041$\pm$0.005} & \underline{0.819$\pm$0.010} & \underline{0.057$\pm$0.005} & \underline{0.064$\pm$0.010} & \underline{0.064$\pm$0.010} \\
        (ICCV'25) & 60\% & 0.831$\pm$0.004 & \underline{0.054$\pm$0.004} & \underline{0.046$\pm$0.003} & \underline{0.049$\pm$0.003} & 0.758$\pm$0.006 & \underline{0.103$\pm$0.021} & \underline{0.095$\pm$0.001} & \underline{0.096$\pm$0.002} \\ \addlinespace
        
        \textbf{SafeECGMatch} 
         & 30\% & \textbf{0.852$\pm$0.005} & \underline{0.045$\pm$0.015} & \textbf{0.036$\pm$0.003} & \textbf{0.037$\pm$0.007} & \textbf{0.829$\pm$0.009} & \textbf{0.052$\pm$0.014} & \textbf{0.057$\pm$0.003} & \textbf{0.057$\pm$0.003} \\
        (\textbf{Ours}) & 60\% & \textbf{0.843$\pm$0.009} & \textbf{0.047$\pm$0.011} & \textbf{0.043$\pm$0.006} & \textbf{0.042$\pm$0.007} & \underline{0.780$\pm$0.018} & \textbf{0.086$\pm$0.027} & \textbf{0.079$\pm$0.018} & \textbf{0.077$\pm$0.018} \\
        \bottomrule
    \end{tabular}%
    }
\end{table}
\section{Sensitivity Analysis}
\begin{table}[H]
    \centering
    \caption{Sensitivity analysis of $\lambda_{\mathrm{sum}}$ on PTB-XL with 60\% OOD. Each experiment is repeated with three random seeds. Higher ACC is better, while lower ECE, ACE, and SCE indicate better calibration.}
    \label{tab:sensitivity-branch-weight}
    
    \setlength{\tabcolsep}{6pt}
    \renewcommand{\arraystretch}{1.15}
    
    \begin{tabular}{lcccc}
        \toprule
        \textbf{$\lambda_{\mathrm{sum}}$} & \textbf{ACC $\uparrow$} & \textbf{ECE $\downarrow$} & \textbf{ACE $\downarrow$} & \textbf{SCE $\downarrow$} \\
        \midrule
        
        $\lambda_{\mathrm{sum}}=\frac{1}{3}$ 
        & 0.842$\pm$0.009 
        & 0.056$\pm$0.008 
        & 0.046$\pm$0.008 
        & 0.046$\pm$0.008 \\
        
        $\lambda_{\mathrm{sum}}=1$ 
        & 0.843$\pm$0.009 
        & 0.047$\pm$0.011 
        & 0.043$\pm$0.006 
        & 0.042$\pm$0.007 \\
        
        $\lambda_{\mathrm{sum}}=3$ 
        & 0.847$\pm$0.010 
        & 0.048$\pm$0.005 
        & 0.041$\pm$0.002 
        & 0.042$\pm$0.003 \\
        
        \bottomrule
    \end{tabular}
\end{table}
\end{document}